\documentclass[letterpaper, 10 pt, conference]{ieeeconf}  

\IEEEoverridecommandlockouts                              

\usepackage{graphics} 
\usepackage{epsfig} 
\usepackage{amsmath} 
\usepackage{amssymb}  
\usepackage{mathabx}
\usepackage{multirow}

\usepackage[normalem]{ulem}
\usepackage{color}

\title{\LARGE \bf
Deep Workpiece Region Segmentation for Bin Picking
}

\author{Muhammad Usman Khalid$^{1+}$, Janik M. Hager$^{2*}$, Werner Kraus$^{1}$, Marco F. Huber$^{3}$, Marc Toussaint$^{2}$ 
\thanks{$^{+}$ Contributed substantially in implementing real data labelling framework, complete fully convolutional network pipeline and experimentation.}%
\thanks{$^{*}$ Contributed substantially in developing simulated data generation pipeline and first patch-based framework used as baseline method \cite{janik2018icra}.}%
\thanks{$^{1}$Robot and Assistive Systems,
        Fraunhofer IPA, Stuttgart, Germany
        {\tt\small muk@ipa.fraunhofer.de}}%
\thanks{$^{2}$Machine Learning \& Robotics Lab, University of Stuttgart, Germany
        {\tt\small janik.hager@ipvs.uni-stuttgart.de}}%
\thanks{$^{3}$Center of Cyber Cognitive Intelligence (CCI), Fraunhofer IPA, Stuttgart and Institute of Industrial Manufacturing and Management IFF, University of Stuttgart, Germany.}
}

\begin{document}

\maketitle
\thispagestyle{empty}
\pagestyle{empty}

\begin{abstract}

For most industrial bin picking solutions, the pose of a workpiece is localized by matching a CAD model to point cloud obtained from 3D sensor. Distinguishing flat workpieces from bottom of the bin in point cloud imposes challenges in the localization of workpieces that lead to wrong or phantom detections. In this paper, we propose a framework that solves this problem by automatically segmenting workpiece regions from non-workpiece regions in a point cloud data. It is done in real time by applying a fully convolutional neural network trained on both simulated and real data. The real data has been labelled by our novel technique which automatically generates ground truth labels for real point clouds. Along with real time workpiece segmentation, our framework also helps in improving the number of detected workpieces and estimating the correct object poses. Moreover, it decreases the computation time by approximately $1$s due to a reduction of the search space for the object pose estimation.

\end{abstract}


\section{Introduction}
The task of grasping and picking objects belongs to the most important task and is still difficult to perform reliably for robots. Solving this task is fundamental as the ability of grasping an object allows solving more complex tasks relying on it, like emptying bins for industrial applications. Several methods are available to find suitable grasps in a cluttered environment. But the search space for finding such a grasp location is often quite huge since the whole image, either containing RGB, RGB-D or just depth information, has to be taken into account. This slows down the whole process which is crucial for industrial tasks because strict time is a constraint and therefore fast solutions in real time are demanded.

For some cases, end-to-end approaches like \cite{levine2018learning, mahler2017dex} are suitable enough to solve grasping and bin picking tasks. However, for industrial use cases, the pose of the objects is critical to be determined correctly since the objects have to be processed further which is why end-to-end approaches typically fail for these scenarios. Therefore, our approach uses a post-processing method where the pose of the object to be grasped is estimated and a suitable grasp point is computed. The main part of our framework aims for segmenting the objects from the background and getting a generalized network which can segment out any unseen objects from background. Segmenting the background from the image appears to be especially difficult for industrial use cases since some of the scenarios might be more complex, e.g., having flat workpieces, damaged bin walls or extra packaging in the box. In this paper, our work mainly deals with solving the problem of segmenting out flat workpieces from a bin.

\begin{figure}[t!]
  \centering
  \includegraphics[scale=0.75]{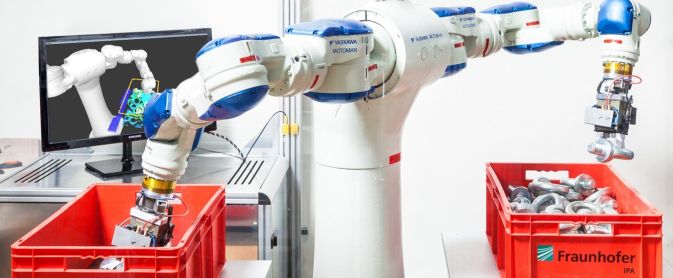}
  \caption{Bin Picking Robot Cell for collecting training data and experiment.}
  \label{FCN_main}
\end{figure}

Depth information is one of the most important ingredient in computing grasps, especially regarding path trajectories and collision checking. Several of the existing grasping methods also use depth information for finding objects and grasp points. Therefore, our work focuses on reducing the search space for object and grasp point localization in depth images. Furthermore, this limits the number of applicable sensor types and data to only having depth sensors, point clouds and the corresponding projections as depth maps.

Overall, our framework, based on a fully convolutional neural network, can reliably segments workpiece regions and background of complex and difficult workpieces and bin situations in a point cloud. To further improve the generalization of our framework, we created a simulation environment to collect training data for different kinds of workpieces and boxes. Our network is trained on this simulated data combined with real data which has been gathered from a robot bin picking cell as shown in Fig. \ref{FCN_main}. The data is automatically labelled by a tool we designed for this task to reduce the processing time and manual correction efforts. In doing so, new training data can easily be created and new workpieces can be easily integrated. Thus, our contribution consists of 1) a framework based on a fully convolutional neural network capable of segmenting arbitrary workpiece regions in challenging scenarios (e.g., flat workpieces) using depth data only and 2) a data labelling tool that allows the framework to be adapted easily for new workpieces and new bin shapes.

\section{Related Work}
Bin picking is considered as a special case of object grasping, as it limits the environment to a box to pick objects from. This task is tackled by different kinds of approaches that can be roughly divided into two main categories: end-to-end methods directly extracting grasp points from the image and methods using an object detection framework beforehand to find suitable grasp points on the detected objects.

The end-to-end methods typically provide a heat map which encodes the probability of a successful grasp. In doing so, the robot has to search for the highest probability to be able to grasp the objects. A problem which might arise with this approach is that the robot might not know the orientation of the grasped object or even which object has been picked at all. On the other hand, these approaches are independent of previously defined objects and grasp points. Recent end-to-end approaches have strongly improved the results of object grasping, e.g., the use of several robots in parallel \cite{levine2018learning} or computing the heat map from artificial grasps in simulations for parallel grippers \cite{mahler2017dex}.

The other class of approaches tries to find suitable grasp locations by first estimating the pose of object and then extracting grasp points. Such a method is being used by the Bin Picking 3D software, described in \cite{dieter2003intelligent, palzkill2012object, spenrath2013object}, which is used by our approach as post-processing step to determine object pose in a point cloud. It searches for local maxima in the depth map to start a pose estimation of the known object by fitting the CAD model and determining its pose using a look-up table. Afterwards a collision free path is computed to a predefined grasp location on the object.

Object detection and semantic segmentation are popular topics and lots of research being done in these topics has improved the achieved results. State-of-the-art results are being pushed by strong frameworks like Fast/Faster/Mask R-CNN \cite{girshick2015fast, ren2015faster, he2017mask}. The use of CNNs to solve semantic segmentation tasks in terms of image classification is quite popular and showed superior results in recent works \cite{wang2018understanding, chen2018encoder, chen2018deeplab}. Especially the usage of fully convolutional networks seems promising \cite{long2015fully, noh2015learning} which is why our framework adapts the same general structure. Some segmentation methods are also specifically configured for bin picking tasks like \cite{hema2007segmentation, schwarz2018rgb, danielczuk2018segmenting}. Other than these methods, our framework considers only object-background segmentation as part of semantic segmentation, i.e., only two classes should be distinguished, namely important objects to interact with in the scene and background. We show that this is especially beneficial for bin picking tasks since the background consists of any environmental observations and the bin while the objects to be grasped can belong to any kind of class. Therefore, our approach is independent from previously chosen object classes and can successfully segment the objects to be picked from the background. Furthermore, most of the approaches make use of RGB data which is why our approach provides a solution using depth maps only to reduce the amount of used data.


   
\begin{figure}[thpb!]
  \centering
  \includegraphics[scale=0.25]{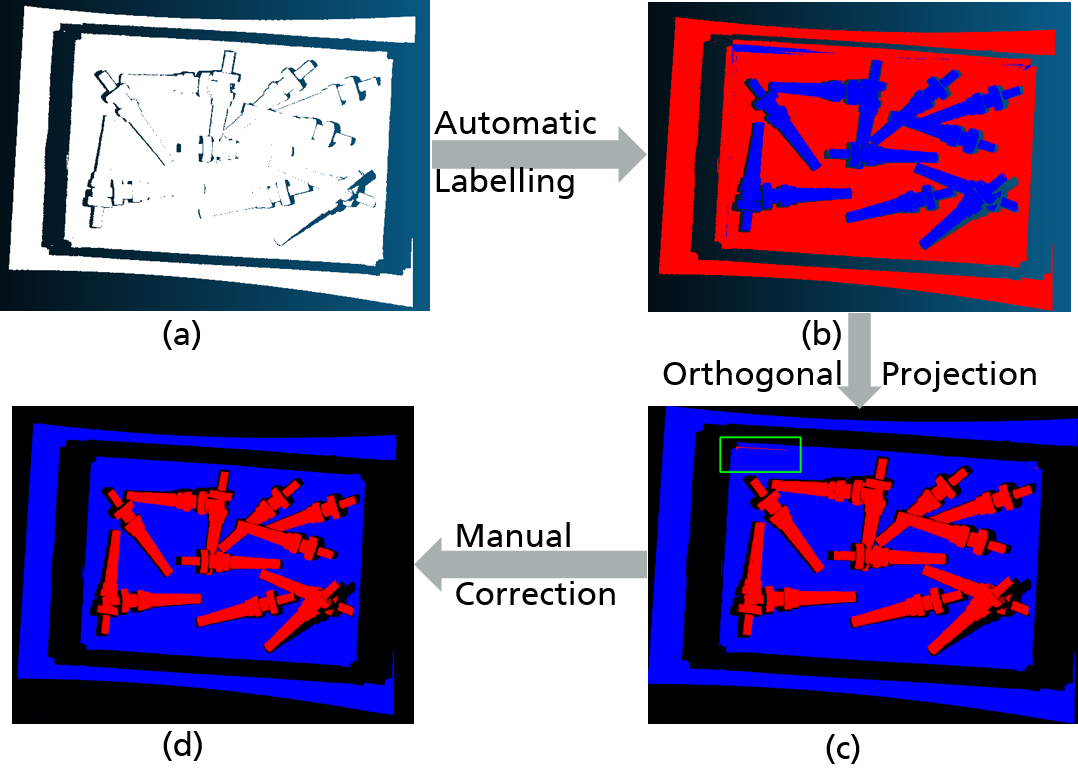}
  \caption{Labelling flow for segmentation. (a) Point cloud $PC_f$ with workpieces, (b) Labelled point cloud $PC_l$ generated by comparison of $PC_f$ and $PC_e$, (c) Orthogonal projected depth map $DM_p$, Green rectangle shows wrongly labelled region (d) Labelled depth map $DM_l$ by manual correction of wrong labels.}
  \label{labelling}
\end{figure}

\section{Data Generation}
For most machine learning algorithms, a large amount of labelled data is required. In our case, we want to use both simulated and real data for training. Regarding simulated data, we create a simulation environment to automatically generate data and the corresponding labels. Labelling of the real dataset requires most of the effort and resources. For a machine learning algorithm to easily adapt to new data, it is always essential to have a process which can do the labelling fast and reliably. We have implemented a framework to do automatic labelling for the workpiece segmentation in point clouds. Additionally, we have also implemented a technique to do manual correction of labels.

\subsection{Simulation environment} \label{sec. simulation environment}
\begin{figure}[t!]
  \centering
  \includegraphics[scale=1.2]{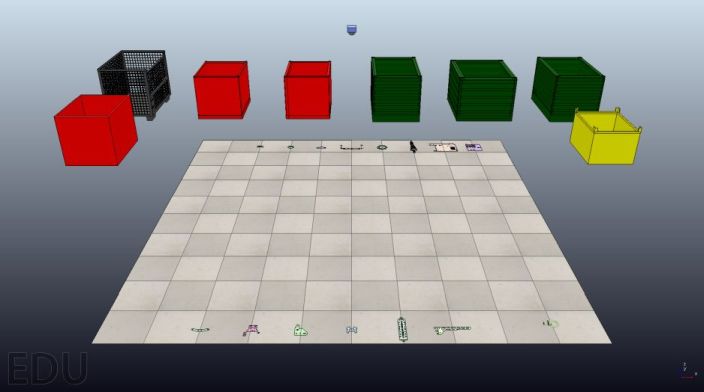}
  \caption{Simulation environment containing the different bin types, workpieces and the sensor.}
  \label{simulation}
\end{figure}

The simulation environment for generating simulated data for training our network is created using Coppelia Robotics' robot simulator V-REP \cite{rohmer2013vrep}. The CAD models of five different box types from industrial use cases are integrated in the scene with two additional variations including small damages. Some of these boxes have structured walls to increase the difficulty, e.g., the lattice box. The boxes are sized from $0.7m \times 0.9m \times 0.6m$ up to $1.0m \times 1.2m \times 0.9m$. Twelve different workpieces, e.g., a gear shaft or a ring bolt are used to fill the bin. Ten of these workpieces are flat, which also increases the difficulty as these are harder to distinguish from the bottom of the box. The whole simulation environment is shown in Fig. \ref{simulation}.

\begin{figure}[t!]
  \centering
  \includegraphics[scale=0.4]{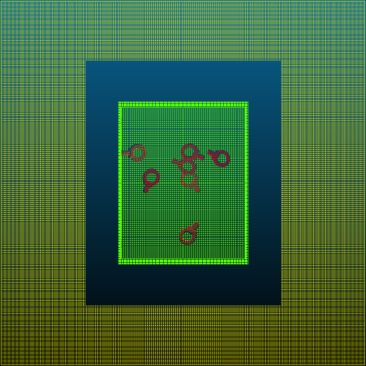}
  \includegraphics[scale=0.4]{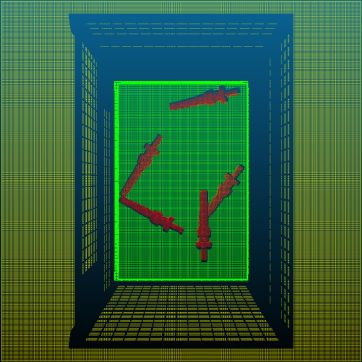}
  \caption{Point clouds created in the simulation environment with two different workpieces and two different bins. Colours correspond to different objects (green = bin, yellow = floor, red = workpieces).}
  \label{simPC}
\end{figure}

For generating data, one of the five boxes is chosen, positioned in the middle of the scene on the planar ground and filled with one type of workpieces. The reason for filling the box with only one type of objects is the close relation with the industrial use cases. While filling the box, the scene is observed by a multifunctional camera collecting depth images, RGB images, point clouds including pixel wise labels for workpiece and environment. This camera is fixed at a height of $2.4m$ above the ground and measures the distances in millimetres. A scan is done every time with random amount of workpieces to increase the variation in collected scenes. The objects are dropped in random orientations and positions above the box. This guarantees a more realistic scenario because the workpieces are also randomly dropped in boxes in the industry as shown in Fig. \ref{simPC}. A full box can contain up to $30$ to $130$ workpieces, depending on the size and shape of the box. Each scanned image has a size of $512 \times 512$ pixels. The data collected in this simulation does not  contain noise of the sensor or perturbations. That makes a huge gap between simulated and realistic data. Therefore, real labelled data is also required so that the neural network can learn more generalized realistic features.

\subsection{Real data labelling}  \label{autolabel}
The tool we developed for labelling real point cloud data for bin picking task consists of several processing steps to generate accurate ground truth data. First of all, for each workpiece and bin scenario, the position of the bin is fixed. Then multiple point cloud scans $PC_e$ of the empty bin are recorded. The inaccuracies in sensor measurements can miss some points in the scene. Therefore, multiple point clouds of the empty bin are recorded to get dense point cloud with no missing point. Keeping the position of the bin fixed, workpieces are added in all kinds of poses and several point cloud scans $PC_f$ of the filled bin are recorded. Using this basic approach, different situations of workpieces and bins can be used to create arbitrary data. In our experimental setting, we use several $3D$ vision sensors, e.g., Stereo vision and laser vision sensors. This ensures the variation of sensor noise in the training data to help in generalization of trained network for segmentation. Next, the point cloud data is mapped to a $X$-$Y$ grid with grid cell size $s_{gc}$. Each point $p_i = (x_i, y_i, z_i)$ of the point cloud is projected to a cell $(x,y)$ as follows:

\begin{equation}
    x = \left\lfloor \frac{x_i}{s_{gc}}  \right\rfloor + 1, \hspace{3mm}
    y = \left\lfloor \frac{y_i}{s_{gc}}  \right\rfloor + 1
\end{equation}
    
Following this formula, all point clouds $PC_e$ with an empty bin for one specific bin and workpiece scenario are mapped to one $X$-$Y$ grid $G_e$. In the same manner, $X$-$Y$ grids $G_{f}$ are created for all point clouds $PC_f$ with a filled bin corresponding to the same bin and workpiece situation.

In the next step, the grid $G_f$ is compared to the grid $G_e$ with the same bin situation to label the point cloud $PC_f$. For each point $p_f \in PC_f$, the euclidean distance $d(p_f, p_e)$ of $p_f$ to all points $p_e \in PC_e$ of the same grid cell $(x,y)$ is calculated and $p_f$ receives a label $l(p_f)$ depending on the calculated distance $d$ as follows:

\begin{equation} \label{eq. auto_label}
    l(p_f) =
    \begin{cases}
    l_w,~~ \text{ if } \forall p_e \in (x,y):~ d(p_f, p_e) > d_{max} \\
    l_n,~~ \text{ if } \exists p_e \in (x,y):~ d(p_f, p_e) \leq d_{max}
    \end{cases}
\end{equation}

This means that if one of these points $p_e$ is closer to $p_f$ than a predefined threshold $d_{max}$, $p_f$ will be labelled as $l_n$ (in our case $l_n = 0$) which means non-workpiece. On the other hand, if none of the points $p_e$ is closer than $d_{max}$, the label is $l_w$ (in our case $l_w = 1$) referring to a workpiece point. This automatic labelling process using eq. (\ref{eq. auto_label}) is done for all point clouds, resulting in fully labelled point clouds $PC_l$. The result of this method is shown in Fig. \ref{labelling}(b). Although creating a nearly perfect labelled point cloud, this automatic labelling process can still generate some wrong labels in the point cloud because of the discretization and noisy sensor data as can be seen in the top left corner of Fig. \ref{labelling}(b). These wrongly labelled points should be corrected manually, as described in the following section.


\subsection{Correction of labels} \label{correctlabel}
To use the labelled point clouds $PC_l$ in our approach, we have to project them orthogonally to create a depth map along with the corresponding labels. The label for each pixel in a depth map $DM_p$ is the label of the highest point being projected at this pixel location. Fig. \ref{labelling}(c) shows the projected depth map along with the labels. To further improve the quality of the ground truth data, we use the morphological closing filter on the generated labels. First some stray non-workpiece labels $l_n$ are removed automatically by applying a dilation filter on the labels with kernel size $k = 3$, i.e. closing holes in the workpiece region. Afterwards an erosion filter is applied with the same kernel size $k = 3$ to keep the initial region sizes. Along with that, it removes stray pixels labelled as workpiece pixels $l_w$. These stray pixel labels are removed automatically to reduce the required amount of manual corrections of wrong labels.

After removing the stray pixel labels, the labelled depth maps are mapped to RGB images where workpiece labels $l_w$ are mapped to red colour and non-workpiece labels $l_n$ are mapped to blue color. Each RGB image is then shown to the user such that user can select the wrongly labelled regions as can be seen by the green rectangle in Fig. \ref{labelling}(c). For each selected region, all the red coloured pixels are changed to blue, simultaneously those pixel labels are changed from workpiece to non-workpiece in the projected depth map $DM_p$. After having manually corrected the labels in the depth map, the correctly labelled ground truth depth map $DM_l$ is saved. Fig. \ref{labelling}(d) shows the finalized labelled depth map $DM_l$. These labelled depth maps can always be projected back to the corresponding point clouds $PC_l$ to correct all wrongly labelled points in them. The remapping to point cloud is explained in Sec. \ref{reprojection}.


\begin{figure}[t!]
  \centering
  \includegraphics[scale=0.24]{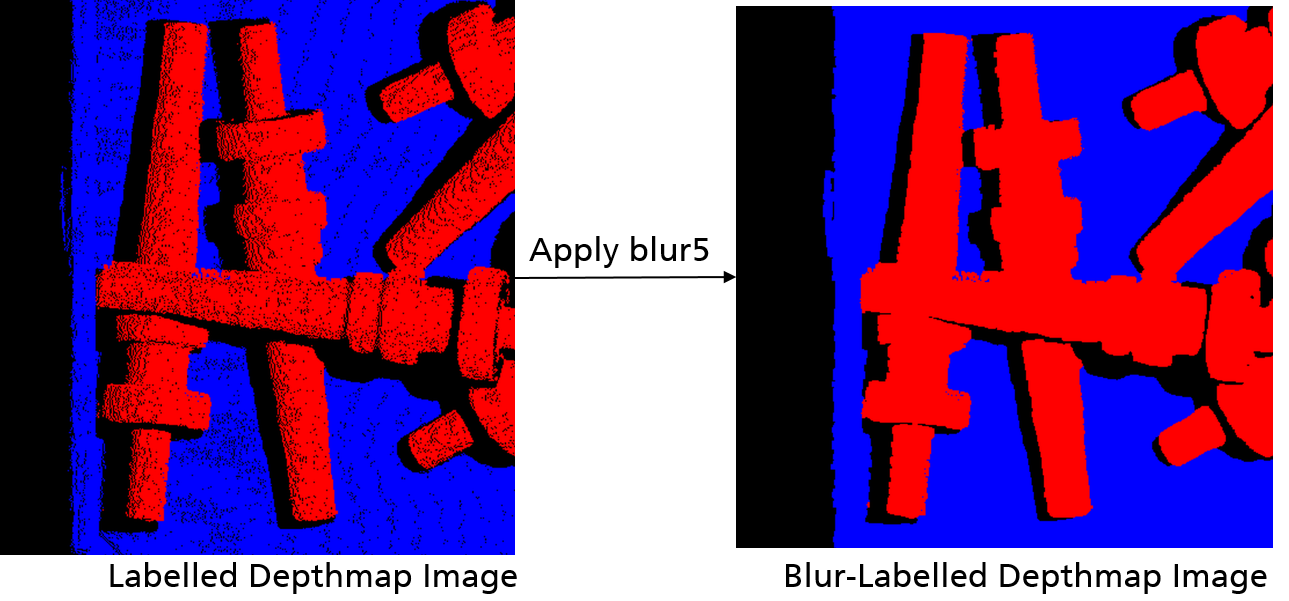}
  \caption{Applying the inpainting method (blur$k$) with $k=5$ to remove holes in the depth map.}
  \label{blurring}
\end{figure}

\begin{figure*}[t!]
  \centering
  \includegraphics[scale=0.39]{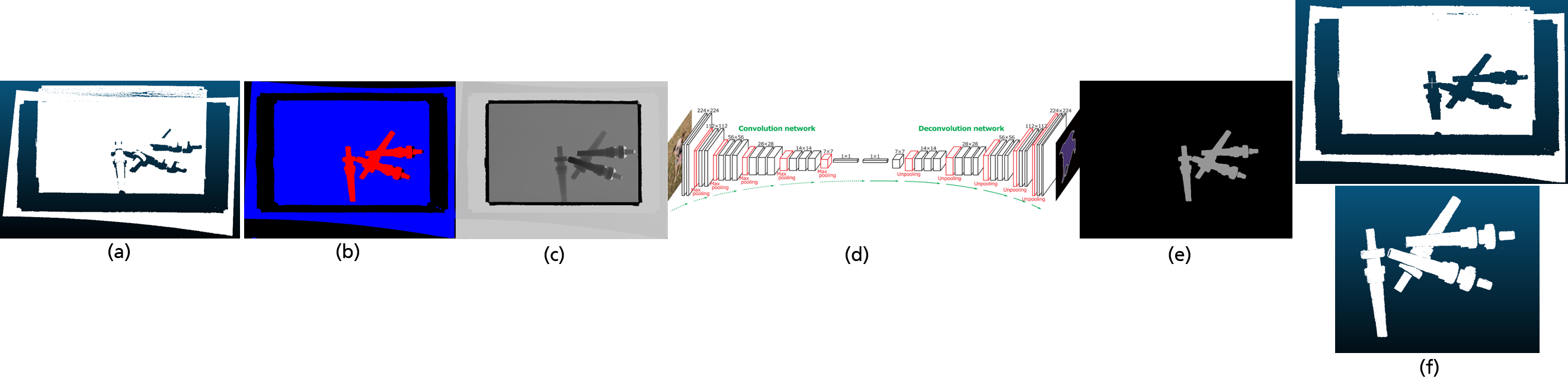}
  \caption{Workflow for workpiece region segmentation on the gear shaft. (a) point cloud $PC$ taken from sensor, (b) orthogonally projected labelled depth map, holes removed and resized to size $s_r$, (c) standarized depth map $DM_r$ used as input for the neural network, (d) fully convolutional neural network \cite{long2015fully} with convolutional and deconvolutional parts, (e) resized segmentation mask $M_r$, (f) projection of segmentation mask to point cloud $PC$ to split it into non-workpiece point cloud $PC_n$ (upper part) and workpiece point cloud $PC_w$ (lower part).}
  \label{workflow}
\end{figure*} 
 
\section{Network Architecture \& Processing Steps}
A fully convolutional neural network \cite{long2015fully} is trained on depth maps with some preprocessing and postprocessing steps. These steps ensure the segmentation of workpiece regions in the point cloud from non-workpiece regions.

\subsection{Projection of point clouds}
Point clouds $PC$ are generated by sensors for each scan. As point clouds are in $3D$ space and the spatial dimension of each point cloud varies, they cannot be fed directly into the neural network. Instead of discretizing the points into a voxel grid, we project the point clouds orthogonally to a $2.5D$ depth map space. A point cloud $PC$ is projected with a resolution factor $r$ to a depth map $DM$ of size $s_{dm} = d_x \times d_y$ with

\begin{equation} \label{eq. pointCloud_projection1}
    d_x = \frac{x_{max} - x_{min}}{r}, \hspace{3mm}
    d_y = \frac{y_{max} - y_{min}}{r}
\end{equation}

where $x_{max}, x_{min}, y_{max}$ and $y_{min}$ correspond to the maximum and minimum values respectively in $x$- and $y$-direction in a point cloud. $r$ determines the resolution factor where $r = 1$ represents each pixel in a depth map $DM$ is mapped to a cell of size $1 \times 1$ in the point cloud. As the point cloud is dense, some points might be mapped to the same cell. In this case, the highest point $p_i = (x_i, y_i, z_i)$ with respect to height $z_i$ is chosen. Fig. \ref{workflow}(c) shows a projected depth map from the point cloud.


\subsection{Hole removal}
Due to shadowing effects and missing sensor data, holes might be created when a point cloud $PC$ is projected to a depth map $DM$ as shown in Fig. \ref{blurring}. These holes can cause ambiguities in training a neural network. We counter this problem with an inpainting method, i.e., the mean value of neighbouring pixels is used to fill the missing pixel values. Only valid pixels contribute to this solution, i.e., either pixels with a value defined by the point cloud or already filled pixels, leading to the following equation:

\begin{equation} \label{eq. hole removal}
    p(x,y) = \frac{1}{n} \sum_{(\tilde{x}, \tilde{y}) \in \mathcal{N}_k(x,y)} p(\tilde{x}, \tilde{y})
\end{equation}

where $\mathcal{N}_k(x,y)$ is the neighbourhood of size $k \times k$ around a pixel at position $(x,y)$ and $n$ is the number of neighbouring pixels with valid values. For $k=3$, this would give us the eight neighbouring pixels around the center pixel. In general, this method has a blurring effect, hence we also refer to it as blur$k$. This hole filling filter is iterated over the whole depth map image to fill all missing pixels. The label value of the filled pixel corresponds to the majority vote of its valid neighbouring pixels. Fig. \ref{blurring} shows the image after applying the hole filling filter with a size of $k = 5$. As can be seen, this filter has filled the black pixels within workpiece and bin regions in a meaningful manner.


\subsection{Resizing of depth maps}
Each projected depth map can have arbitrary image sizes $s_{dm}$ depending on the spatial resolution of the scanned point cloud. However, our fully convolutional neural network can only take inputs of a fixed image size $s_r$. To keep the originality, aspect ratio of the depth map and to preserve the shape characteristics in depth maps, instead of doing interpolation to resize depth maps, an extra zero padding is applied along the spatial dimensions of depth maps for the depthmaps with spatial dimensions being smaller than $s_r$. If the original depth map $DM$ has a spatial dimension greater than $s_r$, cropping is applied to resize the depth map. The resized depth map $DM_r$ is then fed directly to the neural network for training. Usually, point clouds are projected with resolution factor $r=1$, however, if both spatial dimensions of the projected depth map $DM$ are greater than the resized image shape $s_r$, point clouds are projected with greater resolution factor such as $r=2$. It ensures that the regions of interest of the bin and the workpieces are not cropped away while resizing the depth map.
 

\subsection{Training of neural network}
For training a neural network, a segmentation mask $M$ is prepared from the labelled depth map $DM_l$. The segmentation mask is then resized to $M_r$ accordingly to the previous resizing step of the depth map $DM$ to match with the spatial dimensions of its resized version $DM_r$. In this segmentation mask $M_r$, each pixel is either labelled as background class $0$ or workpiece class $1$. The resized depth map $DM_r$ is then normalized by subtracting the mean from all the pixel values and dividing by its standard deviation, such that the depth map has a zero mean and a standard deviation of one. Following the architecture of \cite{long2015fully}, a fully convolutional network is created which can take normalized depth maps as its input and output the corresponding segmentation masks. The network is trained in an end-to-end fashion with per-pixel multinomial logistic loss function. The error is calculated by taking the mean over the loss for both classes. The network is validated by taking the standard metric of mean pixel intersection over union (IoU), with mean taken over both classes, including background. 

The network architecture consists of a convolutional and a deconvolutional network. The convolutional part of the network consists of VGG16 network \cite{long2015fully} where fully connected layers are replaced by convolutional layers. Following \cite{long2015fully}, for the deconvolutional part, FCN-8s architecture is created by fusing predictions from pool3 with a $2 \times$ upsampling of predictions fused from pool4 and conv7. The weights of the VGG16 network are pretrained on ImageNet \cite{deng2009imagenet} while the weights of the FCN-8s network are trained from scratch by initializing them with random values sampled from a Gaussian distribution with zero mean and standard deviation of $0.02$.

\subsection{Resizing of segmentation masks}
The trained neural network predicts a segmentation mask $M_r$ of size $s_r$ for each preprocessed resized depth map $DM_r$. To resize the segmentation mask to the original depth map size $s_{dm}$, unpadding or repadding of zeros has to be done. If additional padding is applied on depth map $DM_r$, the same padding is removed by cropping to readjust the size of the segmentation mask $M$ equal to the size of original depth map $DM$. If cropping is done while adjusting the size of the depth map $DM_r$, an additional zero padding is added on the segmentation masks. This resizing of masks ensures a correct remapping of the segmentation masks to the original point clouds. Such a predicted and resized segmentation mask can be seen in Fig. \ref{workflow}(e).


\subsection{Projection of segmentation masks} \label{reprojection}
After resizing the predicted segmentation masks, each predicted pixel label in the segmentation mask $M$ is mapped back to the corresponding point of the point cloud $PC$, labelling the points either as workpiece point $p_w$ or non-workpiece point $p_n$. This labelled point cloud is then split into two separate point clouds $PC_w$ containing workpiece points only and $PC_n$ containing non-workpiece points only. Fig. \ref{workflow}(f) shows both of the resulting point clouds.

\section{Experimental Results}
Due to unavailability of labelled point cloud data for segmentation and to test the framework in real industrial scenarios, we created our own dataset. This dataset consists of both real data taken from different sensors and also simulated data. We trained several neural networks on different dataset combinations and evaluated the results of workpiece segmentation. Moreover, we also present our cross comparison results of object pose detection with and without segmentation.

\begin{figure*}[thpb]
  \centering
  \includegraphics[scale=0.38]{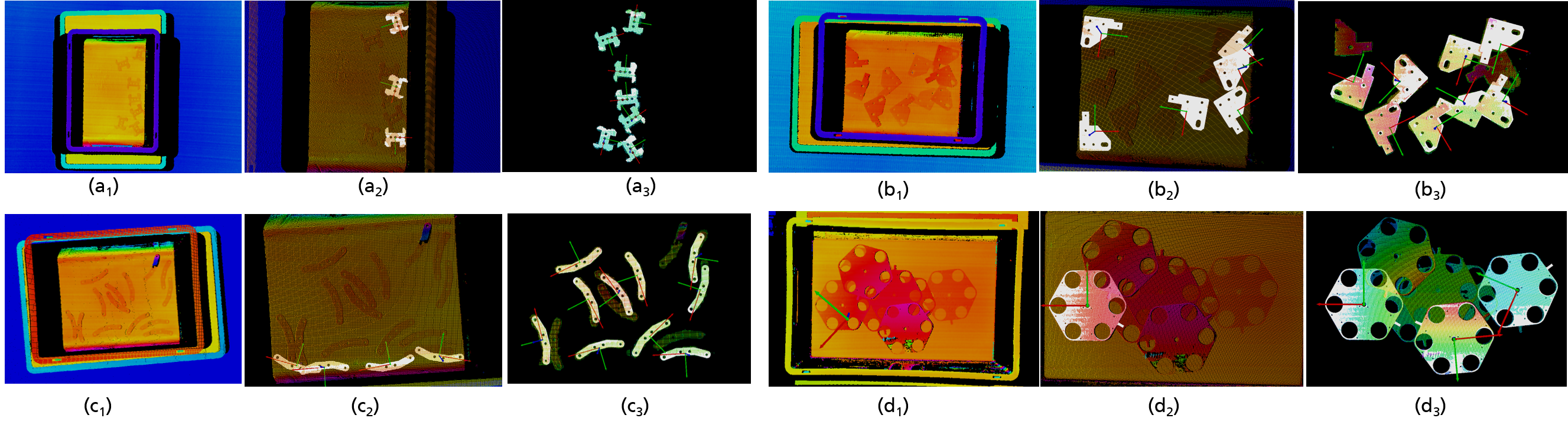}
  \caption{Results for object pose estimation with and without Segmentation. Detected objects are highlighted. Left image ($1$) shows point cloud, center image ($2$) refers to object pose detection without segmentation (often having wrong and phantom detections) and right image ($3$) shows results on point cloud with segmentation (typically correct predictions). ($a$) WP1: $3$ wrong detections vs. $7$ correct detections out of $7$ visible, ($b$) WP2: $6$ wrong detections vs. $8$ correct detections out of $10$ visible, ($c$) WP3: $4$ wrong detections vs. $10$ correct detections out of $14$ visible and ($d$) WP4: $1$ correct detection vs. $3$ correct detections out of $4$ visible.}
  \label{fig. results}
\end{figure*}

\subsection{Description of dataset}
\label{sec. Dataset}
Our dataset is created using twelve different workpieces, of which ten are flat workpieces and two of them have arbitrary geometry. Real scenes are captured with three different sensors, i.e., Ensenso N20, Ensenso X36 and Photoneo Phoxi L. In order to have more variation in our dataset, three different boxes are used, i.e., one box with a size of one quarter of an europalett, one box with a size of one eighth of an europalett and one flat tub with a size of one half of a europalett. Additionally, some data is collected when the workpieces lie on an empty table. All real data is labelled with our automatic labelling along with manual correction tool described in Sec. \ref{autolabel} and \ref{correctlabel}. Overall, we generated 590 labelled depth maps this way. To improve the generalizability of our approach and to further increase the size of our dataset, data augmentation is applied on the depth maps including flipping, scaling and rotation. After data augmentation, the dataset with real data contains 5,150 depth maps. Out of the whole real data, 48 depth maps are reserved for testing, i.e., four depth maps per workpiece. Along with real data, we also generated simulated data with five different bins as described in Sec. \ref{sec. simulation environment}. The amount of simulated data is 2,320 labelled depth maps. For each training experiment with dataset combination, 10\% of randomly chosen data is used for validation.

\subsection{Baseline}
As a baseline for comparison with the proposed framework, we employ a first version of our approach proposed in \cite{janik2018icra}. This baseline is based on a patch-wise CNN to classify the center pixel of each patch while sweeping over the whole depth image.

     \begin{table}[t!]
  \caption{Evaluation of segmentation on real data.}
    \label{table:segmentation}
    \centering
    \begin{tabular}{ |c|c|c| } 
     \hline
     \textbf{Trained Models} & \textbf{Mean Pixel Accuracy} & \textbf{Mean IoU} \\ 
     \hline
     $P$-$CNN_s$ &  79.31 & 29.69 \\
     \hline
     $P$-$CNN_r$ &  97.16 & 70.85 \\ 
     \hline
     $P$-$CNN_{rs}$ &  97.78 &  75.50 \\ 
     \hline
     $FCN_s$ &  85.35 & 48.85 \\
     \hline
     $FCN_r$ &  99.55 & 93.33 \\ 
     \hline
     $FCN_{rs}$ &  \textbf{99.67} &  \textbf{95.78} \\ 
     \hline
    \end{tabular}
    \end{table}
    
\subsection{Evaluation of point cloud segmentation}
To get a generalized neural network for segmenting out workpiece regions and to observe the effect of different data combinations, three fully convolutional neural networks $FCN_r, FCN_s, FCN_{rs}$ are trained and tested with real test data as described in Section \ref{sec. Dataset}. $FCN_r$ is trained only on real data with Adam optimizer and a fixed learning rate of $1e{-5}$ for $350$ epochs (approximately $200,000$ iterations) with a batch size of twelve. Similarly $FCN_s$ and $FCN_{rs}$ are trained on simulated and a combination of real and simulated data respectively with the same training settings as for $FCN_r$. For all of the trained models, the projected depth maps are resized to $s_r=800\times 800$ pixels. The model with lowest validation loss is stored and tested on the testing data. We report mean pixel accuracy and mean intersection over union (IoU) for the workpiece class. The hardware used for segmentation is a NVIDIA DGX-1 with Tesla V100. Table \ref{table:segmentation} shows that the best results are obtained with $FCN_{rs}$ because the trained model has learned better features from real data and the additional simulated data helps for generalization. The model trained on only simulated data $FCN_s$ shows the worst results. The reason for this is that the simulation data is considered as perfect data and the model is unable to learn the perturbations only present in real data. In contrast to this, the $FCN_r$ model shows better results than $FCN_s$ but worse than $FCN_{rs}$ because the real data is limited and the model is not completely generalized. From these results, we can observe that the labelling process of real data and the additional generation of simulated data helps in getting a reliable generalized segmentation network. Moreover, a mean IoU of $95.78\%$ for the $FCN_{rs}$ model also shows the robustness of our approach for segmenting workpiece regions in point clouds.

To benchmark our results, we trained the same three versions of the patch-wise baseline $P$-$CNN_r, P$-$CNN_s, P$-$CNN_{rs}$ \cite{janik2018icra}. Table \ref{table:segmentation} shows that the performance of our fully convolutional approach is clearly superior to the patch-wise framework. Additionally, the patch-wise approach is computationally more expensive as it takes almost 23 sec for segmenting one depthmap compared to 0.08 sec with the fully convolutional approach. Therefore, our approach shows state-of-the art performance and can be used in real time.

\begin{table}[t!]
  \caption{Evaluation of estimated poses in mm and degrees using object pose estimation with and without segmentation.}
    \label{table:OPD_pose}
    \centering
    \begin{tabular}{|c|c|c|c|c|} 
    \hline
     \multirow{2}{*}{\textbf{Workpiece}}	&	\multicolumn{2}{c|}{\textbf{w/o Segmentation}} &	\multicolumn{2}{c|}{\textbf{w/ Segmentation}}\\
    \cline{2-5}
    & pos. err. & orient. err. & pos. err. & orient. err. \\
    \hline
     WP1 & 70.26 & 25.97 & \textbf{1.68} & \textbf{2.67}\\
     WP2 & 121.24 & 47.90 & \textbf{2.43} & \textbf{3.41}\\
     WP3 & 71.68 & 47.32 & \textbf{1.25} & \textbf{0.90}\\
     WP4 & 71.39 & 42.50 & \textbf{6.71} & \textbf{4.67}\\
     WP7 & 162.14 & 52.01 & \textbf{2.27} & \textbf{29.19}\\
     WP10 & 6.22 & 2.35 & \textbf{1.91} & \textbf{2.22}\\
     Ring screw & \textbf{12.91} & \textbf{18.52} & 13.50 & 24.45\\
     Gear shaft & 5.06 & \textbf{4.49} & \textbf{4.75} & 10.55\\
     \hline
     \textbf{Average} & 65.11 & 30.13 & \textbf{4.31} & \textbf{9.76}\\
     \hline
    \end{tabular}
\end{table}

\subsection{Segmentation with object pose estimation}
 In these experiments, an object pose estimation (OPE) technique using generalized Hough transformation \cite{palzkill2012object, spenrath2013object} is applied on the real point clouds. We estimated the ground truth via manually adjusting the CAD model in the scene and then applying the ICP algorithm. Experiments on each workpiece are carried out for ten cycles to estimate object poses with and without our segmentation framework. In these ten cycles, random workpiece situations are created. The $FCN_{rs}$ model is used for segmenting the workpiece region in the point cloud. The OPE parameters are tuned once and the results for estimated poses are recorded. Using estimated and ground truth poses, position and orientation errors are recorded individually. We test the OPE technique on the whole point cloud against the segmented point cloud using our approach. Evidently, as can be seen in Fig. \ref{fig. results}, using our approach, OPE helps to detect more workpieces and especially estimating more correct workpiece poses. Table \ref{table:OPD_pose} presents the mean error of position and rotation estimation for both cases. Here, WP1 to WP10 are flat workpieces and other two have symmetrical geometry. The results show that for all workpieces, the OPE results are improved significantly by segmenting the point cloud first. The segmentation is very effective and helpful for OPE, especially when only a few flat workpieces are present in the bin. It also helps in emptying the bin as last workpieces in bin are difficult to detect. Moreover, we also evaluated the computational time of OPE with segmentation. Depending on the bin and workpiece situation, the OPE without segmentation takes approximately $2-3.5$s in comparison to $1.5-2.5$s of OPE with segmentation on CPU, out of which one forward pass of our neural network took only $80$ms on GPU. The computational time is reduced due to the fact that after the segmentation process, the search space for the object detection is reduced significantly.

\section{Conclusion}
In this paper, we provided a framework to segment workpiece regions reliably from point clouds in real time. We have observed that the workpiece segmentation improves object pose estimation and reduces the search space, hence decreasing the overall computational complexity of object pose estimation. By augmenting the real data and adding simulated data, the trained network generalizes better due to increased variation in the data and hence an improvement in the segmentation results is observed. Another reason for the generalizability of our framework is the usage of different types of bins and workpieces. Therefore, our method is especially helpful for object pose estimation in difficult scenarios, e.g., flat workpieces, which allows completely emptying the bin which was not possible beforehand. Furthermore, with our labelling tool only little effort is required to create new data from point cloud scans and to let our framework adapt to new workpiece situations. In future work, we will further investigate tackling the challenges of additional packaging and we would also like to combine our segmentation method with an object pose estimation approach to create one integrated framework for the whole task.

\addtolength{\textheight}{-12cm}   




\section*{Acknowledgments}

This work was financed by the Baden-W\"urttemberg Stiftung grant NEU016/1 and H2020 project Robott-Net under grant number 688217.


\nocite{*}
\bibliography{bibliography}
\bibliographystyle{ieeetr}

\end{document}